\documentclass[10pt,twocolumn,letterpaper]{article}

\usepackage{authblk}

\usepackage{cvpr}              %

\usepackage{times}
\usepackage{epsfig}
\usepackage{graphicx}
\usepackage{amsmath}
\usepackage{amssymb}
\usepackage{caption}
\usepackage{enumitem}

\usepackage[numbers,sort,compress]{natbib}

\usepackage[pagebackref,breaklinks,colorlinks]{hyperref}

\usepackage{enumitem,amssymb}
\newlist{todolist}{itemize}{2}
\setlist[todolist]{label=$\square$}
\usepackage{pifont}

\def\cvprPaperID{931} %

\def\confName{CVPR}
\def\confYear{2022}

\usepackage{overpic}
\usepackage{enumitem} 
\usepackage{overpic} 
\usepackage{color}

\definecolor{turquoise}{cmyk}{0.65,0,0.1,0.3}
\definecolor{purple}{rgb}{0.65,0,0.65}
\definecolor{dark_green}{rgb}{0, 0.5, 0}
\definecolor{orange}{rgb}{0.8, 0.6, 0.2}
\definecolor{red}{rgb}{0.8, 0.2, 0.2}
\definecolor{darkred}{rgb}{0.6, 0.1, 0.05}
\definecolor{blueish}{rgb}{0.0, 0.3, .6}
\definecolor{light_gray}{rgb}{0.7, 0.7, .7}
\definecolor{pink}{rgb}{1, 0, 1}
\definecolor{greyblue}{rgb}{0.25, 0.25, 1}

\usepackage{blindtext}

\renewcommand{\paragraph}[1]{\vspace{1em}\noindent\textbf{#1}.}
\begin{document}

\definecolor{britishracinggreen}{rgb}{0.0, 0.26, 0.15}
\definecolor{darkspringgreen}{rgb}{0.09, 0.45, 0.27}

\renewcommand*{\Authsep}{, }
\renewcommand*{\Authand}{, }
\renewcommand*{\Authands}{ and }

\newcommand{\beginsupplement}{%
        \setcounter{table}{0}
        \renewcommand{\thetable}{S\arabic{table}}%
        \setcounter{figure}{0}
        \renewcommand{\thefigure}{S\arabic{figure}}%
        \setcounter{section}{0}
        \renewcommand{\thesection}{\Roman{section}}
     }

\title{BARC: Learning to Regress 3D Dog Shape from Images\\by Exploiting Breed Information}

\author[1,2]{Nadine R\"{u}egg} 
\author[3]{Silvia Zuffi} 
\author[1]{Konrad Schindler}
\author[2]{Michael J. Black}

\affil[1]{ETH Z\"{u}rich, Switzerland}
\affil[2]{Max Planck Institute for Intelligent Systems, T\"{u}bingen, Germany}
\affil[3]{IMATI-CNR, Milan, Italy}

\twocolumn[{%
  \renewcommand\twocolumn[1][]{#1}%
  \maketitle
  \vspace*{-1cm}
  \begin{center}
\centerline{  \includegraphics[width=1.0%
    \linewidth]{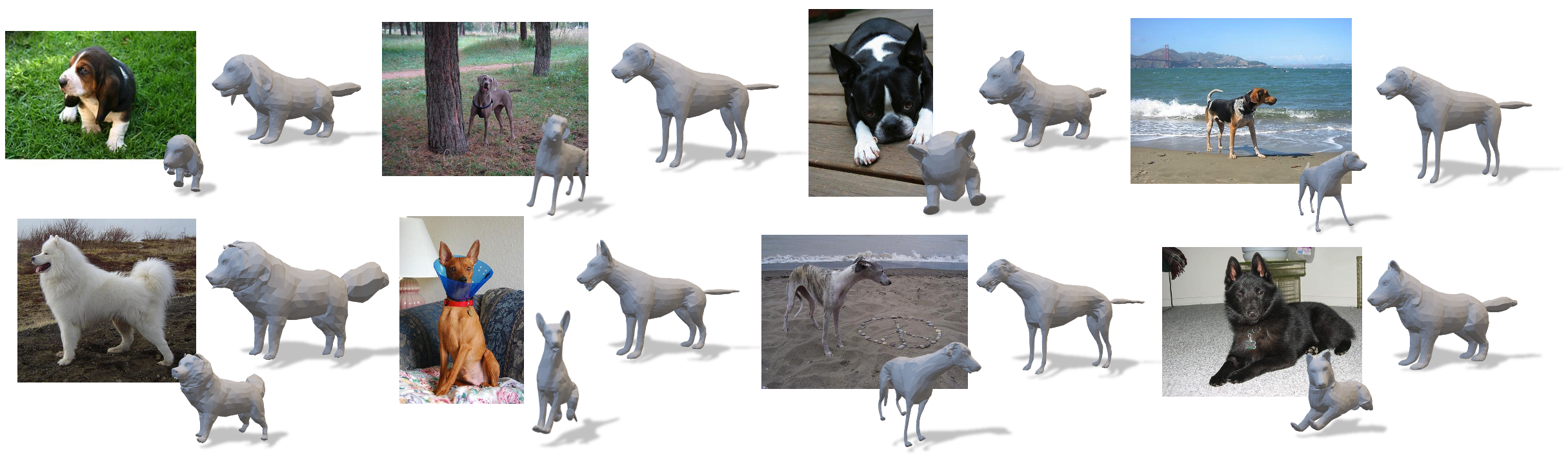}}
    \vspace*{-2.9em}
  \end{center}
  \begin{center}
  \captionof{figure}{\textit{Monocular 3D shape and pose regression of 3D dogs from 2D images.} Since 3D training data is limited, BARC uses {\em breed} information at training time via triplet and classification losses %
  to learn how to regress realistic 3D shapes at test time.
}
  \label{fig:teaser}
  \vspace*{0.3cm}
\end{center}%
}]

\begin{abstract}

Our goal is to recover the 3D shape and pose of dogs from a single image. This is a challenging task because dogs exhibit a wide range of shapes and appearances, and are highly articulated. 
Recent work has proposed to directly regress the SMAL animal model, with additional limb scale parameters, from images.
Our method, called BARC (Breed-Augmented Regression using Classification), goes beyond prior work in several important ways.
First, we modify the SMAL shape space to be more appropriate for representing dog shape.
But, even with a better shape model, the problem of regressing dog shape from an image is still challenging because we lack paired images with 3D ground truth.
To compensate for the lack of paired data, we formulate novel losses that exploit information about dog {\em breeds}.
In particular, we exploit the fact that dogs of the same breed have similar body shapes. 
We formulate a novel {\em breed similarity loss} consisting of two parts: One term encourages the shape of dogs from the same breed to be more similar than dogs of different breeds. The second one, a {\em breed classification loss}, helps to produce recognizable breed-specific shapes.
Through ablation studies, we find that our breed losses significantly improve shape accuracy over a baseline without them. 
We also compare BARC qualitatively to WLDO with a perceptual study and find that our approach produces dogs that are significantly more realistic.  
This work shows that a-priori information about genetic similarity can help to compensate for the lack of 3D training data. This concept may be applicable to other animal species or groups of species.
Our code is publicly available for research purposes at \url{https://barc.is.tue.mpg.de/}.

\end{abstract}

\section{Introduction}

Learning to infer 3D models of articulated and non-rigid objects from 2D images is challenging.  
For the case of humans, recent methods leverage detailed parametric models of human body shape and pose, like SMPL \cite{SMPL_2015}.
Such models have been learned from thousands of high-resolution 3D scans of people in varied poses.
This approach cannot be replicated for most animal species because they are difficult, or even impossible, to scan in a controlled environment. 
Moreover, paired training data of animals with known 3D shape is even rarer.
To make progress, we must leverage side information that can be easily obtained, yet constrains the task of 3D shape and pose estimation.

The 3D reconstruction of animal shape and pose has many real-life applications, ranging from biology and biomechanics to conservation. Specifically, the non-invasive capture of 3D body shape supports morphology and health-from-shape analysis. Markerless motion capture allows 3D motion analysis for animals that it is not possible to capture in a lab setting.        
Here we focus on dogs as a rich, representative, test case.
Dogs exhibit a wide range of shapes, are non-rigid, and have complex articulation. 
Consequently, dogs are challenging and representative of many other animals.

Here, our goal is to learn to estimate a dog's 3D shape and pose from a monocular, uncontrolled image.
Given the lack of 3D training data,  we train a regression network with 2D supervision, in the form of keypoints and silhouettes. 
With only such 2D information, the problem is, however, heavily under-constrained: many 3D shapes can explain the 2D image evidence equally well.
To make the task well-posed, we need additional, prior information.
Here we explore a novel source of a-priori knowledge: a dog's shape is determined, in part, by its {\em breed}. 
Even a trained amateur can recognize the breed by looking at a dog's shape (and appearance). %

\begin{figure}[!t]
\centerline{
  \includegraphics[width=1\linewidth]{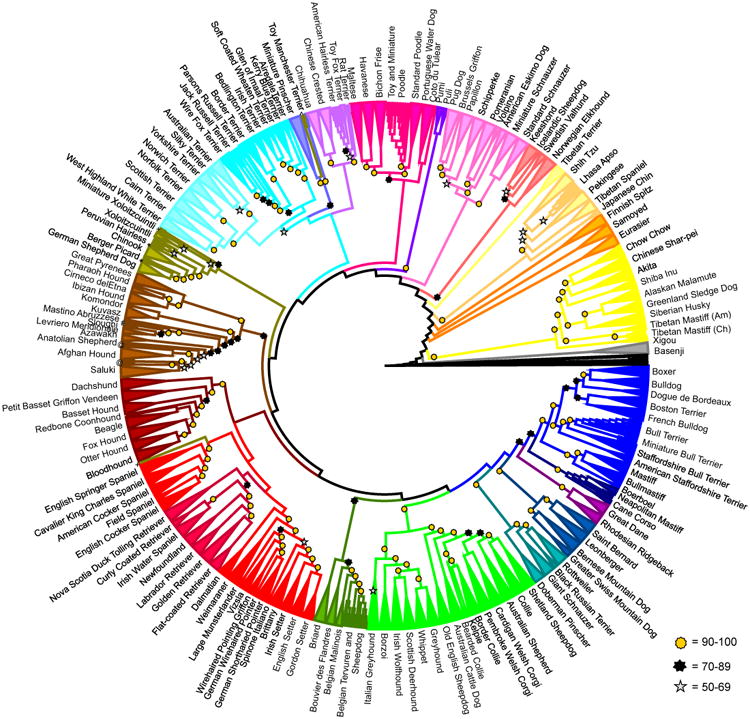}}
  \vspace{-0.1pt}
  \caption{\textit{Cladogram of domestic dog breeds}. The diagram represents clustering according to genetic similarity. Reproduced from~\cite{Parker2017}.}
\label{fig:cladogram}
\end{figure}

Dogs are a particularly interesting case to explore the role of breeds because of their large variety.
Dogs have been domesticated and bred for a long time, for diverse purposes such as companionship, hunting, or herding, but also racing, pulling sleds, finding truffles, etc.
Consequently, breeders have selected for a range of traits including body shape (as well as temperament, appearance, etc.) which has led to a large number of breeds with very different characteristics.
A recent analysis of the dog genome illustrates the relationship between different breeds that exist today \cite{Parker2017}. Breeds are grouped into \emph{clades}, often with high shape similarity within a clade. Figure~\ref{fig:cladogram} shows a cladogram of 161 domestic dog breeds~\cite{Parker2017}.

Here, we explore the use of genetic side information, in the form of breed labels, to train a regressor that infers 3D dog shape from 2D images.
Specifically, we train a novel neural network called {\bf BARC}, for ``Breed-Augmented Regression using Classification."
We follow the approach of regressing a parametric 3D shape model directly from image pixels, which is common in human pose and shape estimation.
Here, we use the SMAL animal model~\cite{zuffi20173d} to define the kinematic chain and mesh template. 
We extend SMAL in several ways to be a better foundation for learning about dog shape, %
this includes adding limb scale factors and extending its shape space with additional 3D dog shapes.

To solve the problem of estimating dog shape from images, we make several contributions.
(1) We propose a novel neural network architecture to regress 3D dog shape and 3D pose from images.
(2) To make training feasible from 2D silhouettes and keypoints, we exploit the fact that 2D images of the \emph{same} breed should produce similar 3D shapes, while different breeds (mostly) have different shapes. With this assumption, we impose classification and triplet losses on the training images, which come with breed labels.
(3) As a result, we learn a \emph{breed-aware latent shape space}, in which we can identify breed clusters and relationships in agreement with the cladogram in Fig.~\ref{fig:cladogram}.
(4) Optionally, we show how to exploit 3D models, if available for some breeds.

Although we use one of the largest dog datasets in the literature, the large number of dog breeds (in our case $120$) means there are only a few images per breed. One can interpret our method as learning a common shape manifold for all dogs (as not enough examples are available per breed), while using the breed labels to locally regularize it.
To our knowledge, this is the first method that exploits breed information to regress the 3D shape of animals from images.

\begin{figure*}[!t]
\centerline{
    \includegraphics[width=1\linewidth]{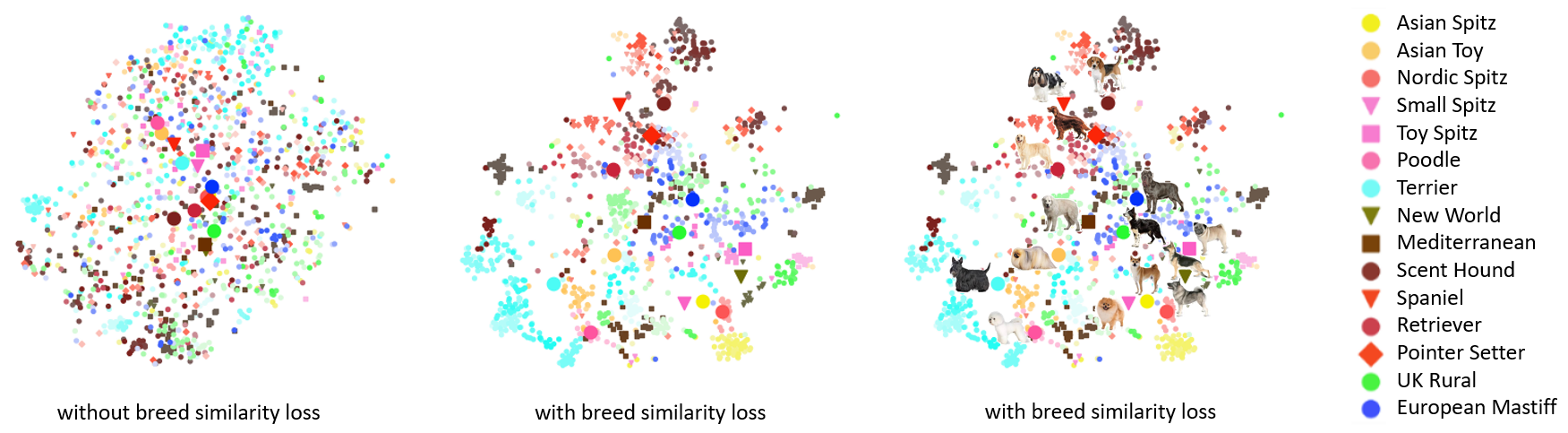} } %
  \vspace{-0.1in}
  \caption{\textit{Learned latent space.} t-SNE \cite{vanDerMaaten2008} visualization of the 64-dimensional latent shape variable for dogs in the test set. Large markers indicate average values within each of the clades in Fig.~\ref{fig:cladogram}. \textit{Left}. Latent space of the network trained without breed similarity loss. Note that the clade means are all near the population mean, indicating poor clustering. \textit{Center and right}: With breed similarity loss. For each clade, colors with different saturation indicate different breeds within the clade.}
\label{fig:latent}
\end{figure*}

We train the network on the Stanford Extra (StanExt)~\cite{biggs2020wldo, KhoslaYaoJayadevaprakashFeiFei_FGVC2011} training set, which we extend with eye, withers and throat keypoints, and test our approach on the StanExt test set. 
We evaluate on this dataset of 120 different dog breeds and show that we learn a latent shape space for dogs in which more closely related dogs are in closer proximity (Fig.~\ref{fig:latent}).
Through ablation studies, we evaluate the impact of different types of breed information and find that each loss leads to a significant improvement in shape accuracy.
We evaluate accuracy with standard 2D measures like PCK and IOU, but these do not necessarily reflect 3D accuracy.
Consequently, we create a dataset of 3D dogs to evaluate shape of corresponding breeds. This allows a quantitative evaluation, and we significantly outperform the prior art (WLDO \cite{biggs2020wldo}). 
Finally, to evaluate shape estimates for in-the-wild images, we use a perceptual study to compare methods.
We find that our final model is more realistic than ablated versions or WLDO.

\section{Related Work}

While many approaches focus on 3D reconstruction of humans from images, there is comparably little work on animal 3D pose and shape estimation.   
Animal reconstruction from images has been approached in two main ways: model-free and model-based.

\noindent \textbf{Model-free 3D Reconstruction}. %
These methods do not exploit an existing  3D shape model. Ntouskos et al.~\cite{Ntouskos2015}  create 3D animal shapes by assembling 3D primitives obtained by fitting manually segmented parts in multiple images of different animals from the same class. Vicente and Agapito~\cite{Vicente2013} deform a template extracted from a reference image to fit a new image using keypoints and the silhouette, without addressing articulation. 
Kanazawa et al.~\cite{kanazawa2018learning} learn to regress 3D bird shape, given keypoints and silhouettes; birds exhibit rather limited articulation. 
Recent work obviates the need for 2D keypoints~\cite{ucmrGoel20,tulsiani2020imr,wu2021dove}.

\noindent \textbf{Model-based 3D Reconstruction}.
In one of the first 3D animal reconstruction methods from images, Cashman and Fitzgibbon \cite{Cashman:2013} deform a 3D dolphin template, learning a low-dimensional deformation model from hand-clicked keypoints and manual segmentation. They also apply their method to a pigeon and a polar bear. A limitation of this approach is that articulation is not explicitly modeled.
In contrast, Zuffi et al.~introduce
 SMAL \cite{zuffi20173d}, a deformable 3D articulated quadruped animal model. Similar to the widely adopted human body model, SMPL \cite{SMPL_2015}, SMAL represents 3D articulated shapes with a low-dimensional linear shape space. Due to the lack of real 3D animal scans, SMAL is learned from scanned toy figurines of different quadruped species. Since dogs are not well represented by SMAL, Biggs et al.~\cite{biggs2020wldo}  extend the SMAL model by adding scale parameters for limb lengths.
 In \cite{wang2021birds}, an articulated 3D model of birds is defined in terms of limb scale variations and used to learn shape from images; it is unclear whether this method easily extends to more complex animals. 
 
The early work using SMAL uses an optimization-based approach to fit the model to image evidence \cite{zuffi20173d} and to refine the animal shapes \cite{zuffi2018lions}.
In other methods, Biggs et al.~\cite{biggs2018creatures} show how to extract accurate animal shape and pose from videos, while Kearney et al.~\cite{kearney2020rgbd} estimate dog shape and pose from RGBD-images.
More relevant to BARC are learning-based methods that regress animal pose and shape directly.
Biggs et al.~\cite{biggs2020wldo} estimate dog pose and shape from single images by regressing pose and shape parameters of their model to training images of the StanExt dataset. Their initial pose prior is improved using expectation maximization with respect to fits of their model to the images. 
Zuffi et al.~\cite{zuffi2019three} regress a zebra SMAL model from images by exploiting a texture map and learn a shape space for the Grevy's zebra.  They train on synthetic data.
In contrast to these methods, Sanakoyeu et al.~\cite{sanakoyeu2020transferring} neither predict 3D directly from the image nor rely on sparsely annotated keypoints. 
Rather they show how to transfer DensePose from humans to a non-human primate. %
This approach does not recover 3D shape or pose.

\noindent \textbf{Supervision without 3D Ground Truth}.
All 3D approaches rely on certain 2D features such as keypoints, segmentation masks or DensePose annotations as a supervision signal. Sometimes those 2D signals are  used as an intermediate representation before the model is lifted to 3D. 
Mu et al. \cite{mu2020learning} exploit synthetic 3D data to predict 2D keypoints and a coarse body part segmentation map. %
They introduce a new dataset for animal 2D keypoint prediction and show how to transfer knowledge between domains, particularly from seen quadruped species to unseen ones. 
Still other work
\cite{ucmrGoel20, kanazawa2018learning, tulsiani2020imr} encourages similarity between objects of similar shape, with small intra-class variability. 
They neither exploit breed information nor use contrastive learning to construct a structured latent space.

\section{Approach}

The present work explores how known breed information at training time can be leveraged to learn to regress a high-quality 3D model of dogs.
To that end, we combine a parametric dog model with a neural network that maps images to model instances. In the following, we describe the model we use, the network architecture in which it is embedded, and the loss functions used to train the architecture, including the novel breed losses.

\subsection{Dog Model}\label{sec:dog_model}
For the parametric representation of a dog's shape and pose, we employ a variant of SMAL.
We start from 41 scanned animal toy figurines of several different species (already used as part of the original SMAL model) as well as 3D \emph{Unity} canine models in the animal equivalent of the canonical T-pose; i.e., standing with straight legs and tail pointing backwards. 
We purchased the same pack of Unity models\footnote{\url{https://assetstore.unity.com/packages/3d/characters/dog-big-pack-105660}} as was exploited by \cite{biggs2020wldo} to initialize their mixture of Gaussian shape prior and use them to relearn the SMAL shape space for our task.
To that end, we fit a mesh with the same topology as SMAL (and WLDO) to the new dogs, add these to the original SMAL training set and recompute the mean shape and the PCA shape space.
The resulting model differs from the original SMAL in three respects: (1) different input data; (2) reweighting of the inputs such that 50\% of the total weight is assigned to dogs; and (3) rescaling of the meshes such that the torso always has length 1.
We further adapt an idea from WLDO and extend the model with scaling parameters $\kappa$ (where the actual scale is $\exp{(\kappa)}$) for the limbs, plus an additional scale for the head length. The scaling is applied to the bone lengths, and propagated to the surface mesh via their corresponding linear blend skinning (LBS) weights.
For compactness, we collect the PCA shape coefficients $\beta_{pca}$ and limb scales $\kappa$ into a shape vector $\beta$.
\subsection{Architecture}\label{sec:architecture}
\begin{figure*}[ht]
\centering
  \centering
  \includegraphics[width=0.8\linewidth]{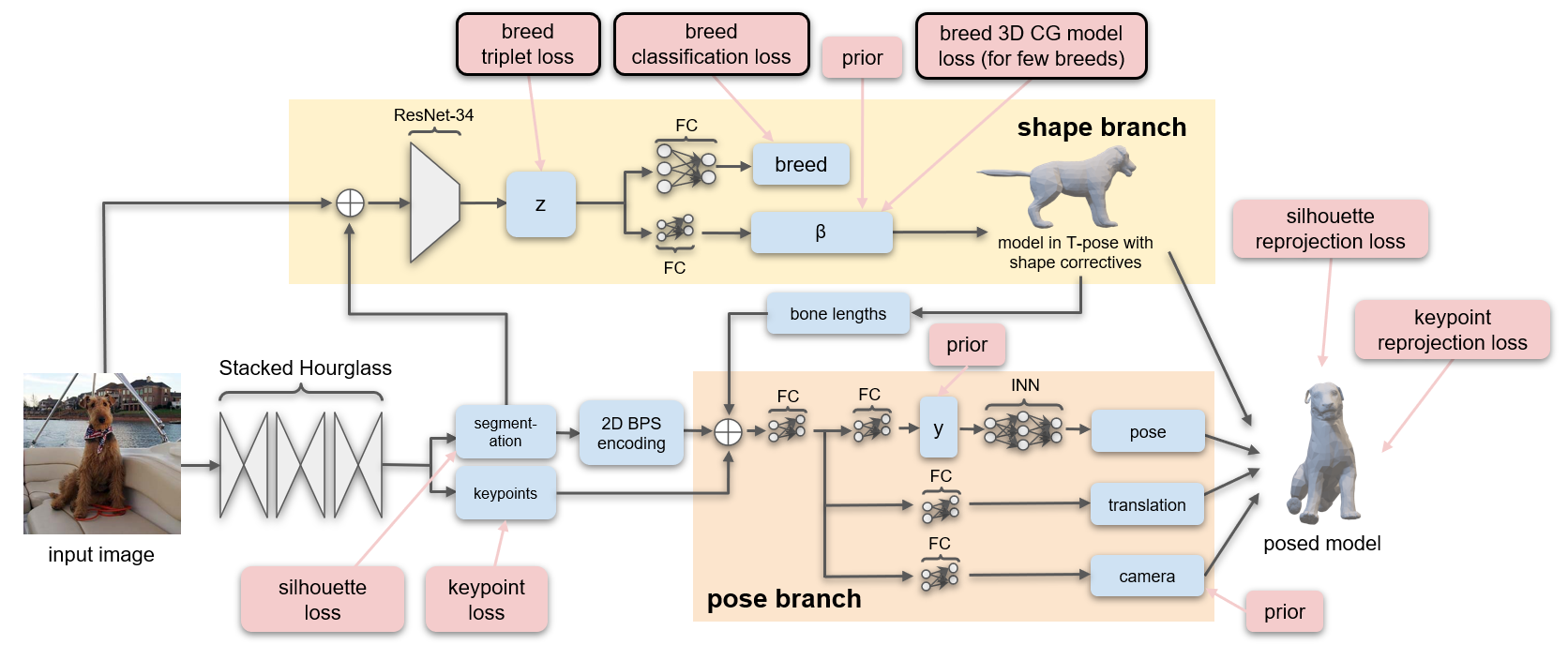} %
  \vspace{-0.1in}
  \caption{\textit{BARC Architecture.} The model consists of a stacked hour glass network followed by two separate branches for shape and pose prediction. Pink boxes illustrate where losses are applied. The pink boxes with black boundaries are our new breed losses.}
\label{fig:architecture}
\end{figure*}

\hspace{-1.4mm} %
Similar to \cite{pavlakos2018learning, zhang2020learning}, we use separate shape and pose branches. Figure \ref{fig:architecture} shows the overall architecture of BARC, consisting of a joint stacked hourglass encoder, a shape branch, a pose branch, and a 3D prediction and reprojection module.

\noindent\textbf{Stacked Hourglass:} First, the input image is encoded and 2D keypoint heatmaps, as well as a segmentation map, are predicted with a pre-trained stacked hourglass network. 2D keypoint locations are extracted from the heatmaps with ``numerical coordinate regression" (NCR,~\cite{nibali2018numerical}). The segmentation map is encoded with a scheme similar to ``basis point sets" (BPS,~\cite{prokudin2019efficient}) for 3D point cloud encoding. To our knowledge, we are the first to apply BSP in 2D. 
Compared to the full segmentation map, this
encoding is lightweight, easy to compute for
silhouettes, and has a similar format as the NCR keypoints. We find that, despite the reduction to a small number of sample points, the silhouette encoding still improves the 3D prediction over 2D keypoints alone. 

\noindent\textbf{Shape-branch:} The input image and the predicted segmentation map are concatenated and fed to a ResNet34 that predicts a latent encoding $z$ of the dog's shape. $z$ is decoded into both a breed (class) score and a vector of body shape coefficients $\beta$.  
We have experimented with different sub-networks between $z$ and $\beta$ and find that the breed similarity loss is most effective when the connection is as direct as possible, with only single, fully-connected layers between $z$ and each of the shape vectors $\kappa$ and $\beta_\text{pca}$. These shape coefficients are applied to the 3D dog template to obtain a shape, whose bone lengths are passed on to the pose branch.

\noindent\textbf{Pose-branch:} The predicted 2D keypoints, the BPS encoding of the silhouette and the bone lengths
from 
the shape network form the 
input
to estimate the dog's 3D pose, its translation \wrt the camera coordinate system and the camera's
focal length.
The pose is represented as a 6D rotation~\cite{zhou2019continuity} for each joint, including a root rotation. Instead of predicting all rotations directly, we predict root rotation and
a latent pose representation $y$. Following recent work on human pose estimation, we implement an invertible neural network (INN) that maps each latent variable $y$ to a pose. This INN is used in the context of a normalizing flow pose prior trained on the RGBD-Dog dataset \cite{kearney2020rgbd}. Similar to \cite{zanfir2020weakly}, we build this network consisting of Real-NVP blocks, but because of  the smaller size of the RGBD-Dog dataset, as compared to AMASS \cite{mahmood2019amass}, our network is much smaller than in previous work
on human pose estimation. 
The aim of the INN is to map the distribution of 3D dog poses to a simple and tractable density function, i.e.~a spherical multivariate Gaussian distribution. 
To train the pose prior, we exploit the RGBD-Dog dataset \cite{kearney2020rgbd}, which contains walking, trotting and jumping sequences, but no sitting or lying poses. 
Note that the INN is pretrained to serve as a pose prior and kept fixed during final network training.

\noindent\textbf{3D Prediction and Reprojection Module:} As a last step, BARC poses the model according to the predicted shape, pose and translation, and reprojects  the keypoints and silhouette to image space, using the predicted focal length.
To minimize the silhouette and keypoint reprojection errors, we employ the Pytorch3D differentiable renderer~\cite{ravi2020pytorch3d}.

\subsection{Training Procedure}
The complexity of articulated, deformable 3D model fitting requires a number of different loss functions, as well as careful pretraining.  %

\noindent\textbf{Stacked Hourglass Pretraining:} The stacked hourglass is pretrained to predict keypoints and the segmentation map. The StanExt dog dataset~\cite{biggs2020wldo} provides labels for both. The keypoint loss consists of two parts, a mean squared error (MSE) between the predicted and true heatmaps, and an L2-distance between the predicted and true keypoint coordinates.
For the silhouette, we use the cross-entropy between ground truth and predicted masks. As usual for stacked hourglasses, we calculate the losses after every stage. 

\noindent\textbf{Pose-branch Pretraining:} 
We use the same dataset (RGBD-Dog) that is used to train the pose prior to also pretrain the pose branch. We sample poses and random shapes and project them to a 256$\times$256 image with a random translation and focal length. 
The projected keypoints and silhouette serve as input to the network. MSE losses are used to penalize deviations between the predicted values and the ground truth. In addition, we use an MSE error between the predicted pose latent representation $y$ and its ground truth.

\noindent\textbf{Main Training:} The stacked hourglass is kept fixed, while all other network parameters are jointly optimized. We point out that we do not have access to 3D ground truth, and, based on 2D keypoints, the true shape and pose is ambiguous. 
To regularize the solution, we therefore combine reprojection losses with suitable priors. These loss terms are described below.

\subsection{Standard Losses}

\noindent\textbf{Keypoint Reprojection Loss} $L^\text{kp}$ is the weighted mean squared error between predicted $k^\text{pred}_n$ and ground truth 2D keypoint locations $k^\text{gt}_n$:
    \begin{equation}
        L^\text{kp} = (\sum_{n=1}^{N_\text{kp}} w_n  d(k^\text{pred}_n, k^\text{gt}_n)^2) / (\sum_{n=1}^{N_\text{kp}} w_n),
    \end{equation}
    where $d(k^\text{pred}_n, k^\text{gt}_n)$ is the 2D Euclidean distance between the predicted and ground truth location of the $n$-th keypoint.
    The weights, $w_n$, balance the influence of keypoints; see Sup.~Mat.

\noindent\textbf{Silhouette Reprojection Loss} $L^\text{sil}$ is the squared pixel error between the rendered $s^\text{pred}$ and ground truth silhouette $s^\text{gt}$:
    \begin{equation}
        L^\text{sil} =
        \begin{cases}
            \sum_{x=1}^{256} \sum_{y=1}^{256} (s^\text{pred}_{xy} - s^\text{gt}_{xy})^2& L^\text{kp,m} < T \\
            0& \text{otherwise.}
        \end{cases}
    \end{equation}
This is used only for images where the mean keypoint reprojection error $L^\text{kp,m}$ is below a threshold $T$. 
    
\noindent\textbf{Shape Prior}: This is a weighted sum of two parts, $L^\text{sh}=w_\beta L^\text{sh}_\beta+w_\kappa L^\text{sh}_\kappa$. The first penalises deviations from a multivariate Gaussian with mean $\mu_{\text{pca}}$ and covariance $\Sigma_{\text{pca}}$:  
    \begin{equation}
        L^\text{sh}_\beta =  (\beta_\text{pca}-\mu_{\text{pca}})^\top \Sigma_{\text{pca}}^{-1} (\beta_\text{pca}-\mu_{\text{pca}}).
    \end{equation}
    Additionally, we penalise deviations from scale 1 with an element-wise squared loss on the scale factors $\kappa$,
    \begin{equation}
        L^\text{sh}_{\kappa} = \sum_{i=1}^{7}\kappa_i^2.
    \end{equation}
    The shape prior loss is assigned a low weight and serves only to stabilise the shape against missing evidence. 

  \noindent\textbf{Pose Prior}: $L^\text{p}$ penalises 3D poses that have low likelihood. Again, it consists of two terms, a normalizing flow pose prior as well as a regularization regarding lateral leg movements. The normalizing flow pose prior penalizes the negative log-likelihood of a given pose sample. 
 Since the learned latent representation $y$ follows a multivariate normal distribution, the pose prior reduces to:
      \begin{equation}
        L^\text{p}_\text{nf} \propto  y^\top y.
    \end{equation}
    The normalizing flow prior is trained on the RGBD-Dog dataset which has a limited set of poses compared to the natural poses in the StanExt dataset.
    Consequently, with only this prior, the network can infer 3D poses where the legs move unnaturally sideways. Thus, we add a second term  $L^\text{p}_\text{side}$ that penalizes sideways poses of the joints in each leg. %
 The final pose prior is:
    \begin{equation}
        L^\text{p} = w_\text{nf} L^\text{p}_\text{nf} + w_\text{side} L^\text{p}_\text{side}\;,
    \end{equation}
with weights $w_\text{nf}$ and $w_\text{side}$, the latter set to a low value.
    
\noindent\textbf{Camera Prior} $L^\text{cam}$: Since focal length $f^\text{pred}$ is heavily correlated with depth (object-to-camera distance), we find it useful to penalise the squared deviation from a reasonably predefined target focal length $f^\text{target}$: 
    \begin{equation}
        L^\text{cam} = (f^\text{pred}-f^\text{target})^2 .
    \end{equation}

\subsection{Novel Breed Losses}

The losses described so far do not depend on the breed. To exploit breed labels for the training images, we introduce an additional breed triplet loss, as well as an auxiliary breed classification loss. We summarize those two losses as breed similarity loss. 
Given the dog meshes used during 3D model learning (Sec.~\ref{sec:dog_model}) we moreover define a specific shape prior for those particular breeds.

\noindent \textbf{Breed Triplet Loss}  $L^\text{B}_\text{triplet}$: Dogs of the same breed usually are somewhat similar in shape. However, this does not imply that there is no intra-class variation, nor that different breeds necessarily have dissimilar shape. Hence, we implement this with a triplet loss. 
We have experimented with different metric learning losses, but found that they all exhibit similar behaviour.
Triplet losses are commonly used in person re-identification (ReID) methods, where the goal is to learn features that are discriminative for person identity \cite{SchroffKP15,taigman2014deepface}.
RingNet used a similar idea to learn 3D head shape from images without 3D supervision \cite{RingNet:CVPR:2019}.

Applying the loss directly to the shape $\beta$ does not work well. Shape changes along different principal directions may have different scales, moreover shape changes due to limb scaling are not orthogonal to the PCA coefficients $\beta_\text{pca}$.
We find it better to apply the triplet loss to the latent encodings $z$. 
Given a batch with an anchor sample $z_{a}$, a positive sample $z_{p}$ of the same breed and a negative sample $z_{n}$ from a different breed, we calculate the triplet loss, $ L^\text{B}_\text{triplet} = $  
    \begin{equation}
       \sum_{i=1}^{N_\text{triplets}} \max (d(z_{a,i},z_{p,i}) - d(z_{n,i},z_{a,i}) + m, 0)\;,
    \end{equation}
where $m$ is the margin and $d$ denotes the distance between the two samples.  

\noindent\textbf{Breed Classification Loss} $ L^\text{B}_\text{cs}$:
    We further bias the estimation towards recognisable, breed-specific shapes with an auxiliary breed classification task, supervised with a standard cross-entropy loss on the breed labels:
    \begin{equation}
        L^\text{B}_\text{cs} = -\sum_{c=1}^{N_\text{classes}}y_{o,c}\log(p_{o,c})\;,
    \end{equation}
    where $p_{o,c}$ is the predicted probability that observation $o$ is of class $c$ and $y$ is a binary indicator if label $c$ is the correct class for observation $o$. 
The full similarity loss reads:
    \begin{equation}
        L^\text{B}_\text{sim} = w_\text{triplet} L^\text{B}_\text{triplet} + w_\text{cs} L^\text{B}_\text{cs}\;,
    \end{equation}
    where $w_\text{triplet}$ and $w_\text{cs}$ are weights. %
    
\noindent\textbf{3D Model Loss}  $L^\text{B}_\text{3D}$:
We have access to a small number of 3D dogs (\emph{Unity} models) and a few 3D scans of toy figurines.
These models encompass 11 of the 120 breeds in StanExt. 
For these breeds, we optionally enforce similarity between the prediction and the available 3D ground truth shape, via a component-wise loss on the shape coefficients $\beta$:
    \begin{equation}
       L^\text{B}_\text{3D} =(\beta^\text{pred}_\text{pca}-\beta^\text{breed}_\text{pca})^2 + (\kappa^\text{pred}-\kappa^\text{breed})^2.
    \end{equation}

\section{Experiments}

We evaluate our approach on the Stanford Extra Dog dataset (StanExt) \cite{biggs2020wldo}. %
StanExt provides labels for 20 keypoints, silhouette annotations and dog breed labels. We extend the 20 keypoints in the training set with withers, throat and eyes. These predictions are obtained by training a separate stacked hourglass on the Animal Pose dataset \cite{cao2019cross}.
Note that this version includes updated results w.r.t. \cite{barc_cvpr22} due to a minor bug fix in the implementation of $L^p_{side}$.

\subsection{Evaluation Methods}\label{sec:evaluation_methods}
\noindent\textbf{2D Reprojection Error: }In the absence of 3D ground truth, it is common to evaluate 3D shape and pose predictions in terms of reprojection errors in image space. We provide results for intersection over union (IoU)  on the silhouette as well as percentage of correct keypoints (PCK). 

\noindent\textbf{Perceptual Shape Evaluation: } 
Many implausible 3D shapes have low 2D reprojection errors, but for in-the-wild images we do not have access to ground-truth 3D shapes that would allow a meaningful comparison.
Instead, we run a study to evaluate relative perceptual correctness where humans visually assess the 3D shapes regressed from in-the-wild images.
Using Amazon Mechanical Turk (AMT), qualified workers judge which of two rendered 3D body shapes better corresponds to a query dog image. 
To focus the workers on shape, we render the dogs in the T-pose.
For an example and details of the task see Sup.~Mat.

\noindent\textbf{Breed Prototype Consistency: } 
Quantitative evaluation of 3D error for uncontrolled images is challenging.
To address this, we exploit the fact that dogs of the same breed have similar shapes.
We define prototype shapes for several breeds with the help of toy figurines that are scanned, registered to the SMAL template, and reposed to the canonical T-pose.
Then, for all StanExt images of the corresponding breeds, we regress their shape using various methods. These predictions are then also transferred to T-pose and aligned to the matching prototype with the Procrustes method.
The vertex-to-vertex error and variance between the estimate and the prototype serve as indicators of how well a given prediction method captures the breed shape.

\subsection{Comparison to Baselines}\label{sec: comparison_baselines}
\begin{figure}[!tb]
    \centerline{    %
    \includegraphics[width=0.98\linewidth]{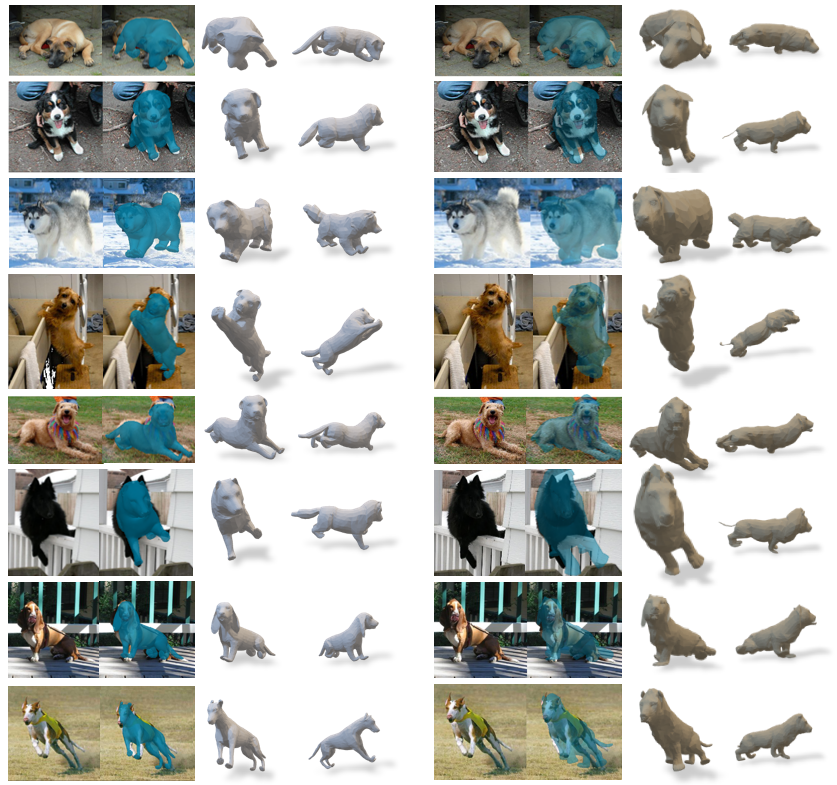}}
\vspace{-0.1in}
    \caption{\textit{Comparison to SOTA}. Qualitative comparison of BARC (left half) with WLDO \cite{biggs2020wldo} (right half). For each method we show input image, the 3D reconstruction projected on the input image, the 3D reconstruction, and a 90$^{\circ}$ rotated view.}
    \label{fig:wldo_comparison}
\end{figure}

\begin{figure*}[!tb]
\centerline{  %
\includegraphics[width=1\linewidth]{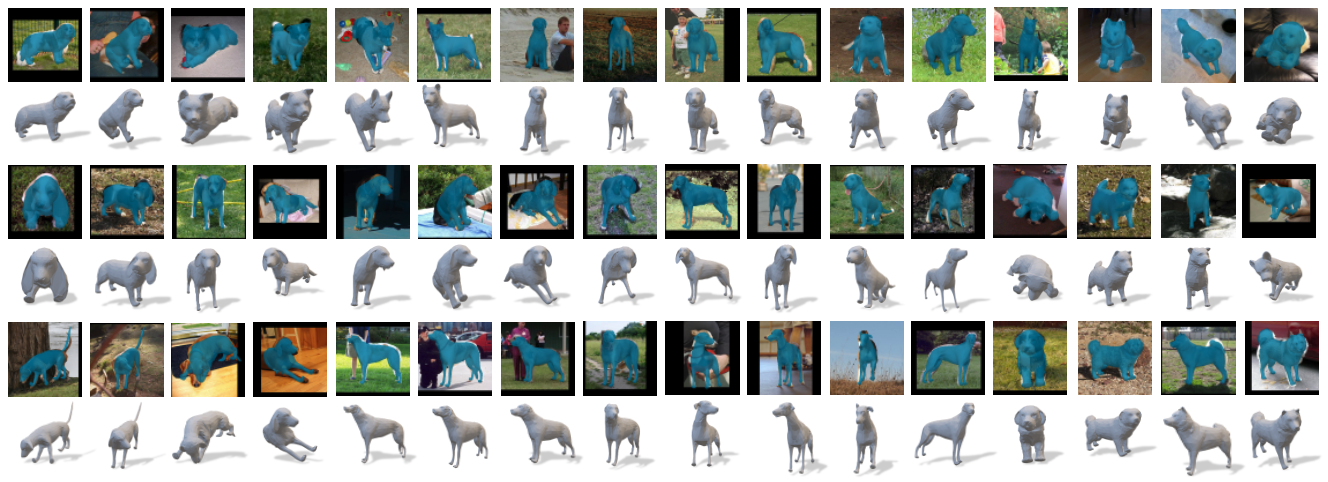}}
\vspace{-0.1in}
  \caption{\textit{BARC results.} Each row shows the input image with the projected 3D shape.  Below that is a rendering of the posed 3D shape. 
  }
\label{fig:results02}
\end{figure*}

In terms of 2D error metrics (IoU and PCK) BARC outperforms prior art, i.e., WLDO \cite{biggs2020wldo}, CGAS \cite{biggs2018creatures} and 3D-M \cite{zuffi20173d}.
Tab. \ref{tab:eval} summarizes the results.
In the perceptual comparison, BARC is also judged to represent the depicted dog better than its closest competitor WLDO, in an overwhelming 92.4\% of all cases. See last line of Tab. \ref{tab:main_amt_breed_losses_main_paper}. 
The marked gap in visual realism is evident in Fig.~\ref{fig:wldo_comparison}. More BARC results, for different breeds, are displayed in Fig.~\ref{fig:results02}.

\begin{table}[!tb]\centering
\centering
\begin{tabular}{|p{1.05cm}||p{0.55cm}|p{0.55cm}|p{0.55cm}|p{0.55cm}|p{0.55cm}|p{0.55cm}|  }
 \hline
 Method  & IoU   & \multicolumn{5}{|c|}{PCK @ 0.15} \\
     &    & Avg  &  Legs  &  Tail  & Ears  &  Face \\
 \hline
  3D-M  & 69.9 & 69.7 & 68.3 & \textbf{68.0} & 57.8 & \textbf{93.7}\\ 
 CGAS  & 63.5 & 28.6 & 30.7 & 34.5 & 25.9 & 24.1\\ 
 WLDO  & 74.2 & 78.8 & 76.4 & 63.9 & 78.1 & 92.1\\  
\textbf{Ours}   & \textbf{75.1} & \textbf{82.8} & \textbf{82.3} & 63.3 & \textbf{83.3} & 91.3\\  %
 \hline
 \end{tabular}
 \vspace{-0.1in}
  \caption{\textit{Comparison to SOTA.} Numbers for
  3D-M \cite{zuffi20173d}, CGAS \cite{biggs2018creatures}, WLDO \cite{biggs2020wldo} reproduced from~\cite{biggs2020wldo}.}
   \label{tab:eval}
\end{table}

\subsection{Ablation Study}
Our key contribution is the addition of breed losses to improve 3D shape regression.
To ablate the impact of individual loss terms, 2D errors are not meaningful, so we again report results in terms of relative perceptual correctness (Tab.~\ref{tab:main_amt_breed_losses_main_paper}) and in terms of consistency with the prototype breed shape (Tab.~\ref{tab:quant_toy_eval}).
We compare three versions of our method:
(i)  our network, trained without any breed losses;
(ii) the same network with the breed similarity losses $L^\text{B}_\text{sim}$, i.e., classification and triplet loss; 
(iii) with all breed losses, including the 3D model loss $L^\text{B}_\text{3D}$.
The results, on both metrics, show consistent improvements with the addition of each breed loss. In terms of perceptual agreement, the two parts of our loss have similar impact.
The triplet and classification losses bring a clear improvement, even though they do not explicitly constrain 3D shape.
Breed-specific 3D shape information can further improve the prediction, but may be difficult to collect at large scale. Note that adding 3D CG models as additional supervision leads to a small improvement (on average) across \emph{all} breeds, even though they are only available for 11 out of 120 breeds. All differences in votes are highly significant ($\chi^2$-test, $p\!<\!0.0001$).

We complement the perceptual study with a quantitative evaluation \wrt breed prototypes (Tab. \ref{tab:quant_toy_eval}). For 20 different breeds we evaluate WLDO, as well as our method without any breed losses, with only $L^\text{B}_\text{sim}$, and with both $L^\text{B}_\text{sim}$ and $L^\text{B}_\text{3D}$.
Already without breed information, our model outperforms WLDO by a clear margin in terms of 3D error, presumably due to technical choices like details of the dog model and network architecture, and the new pose prior. Adding the breed similarity losses decreases the error further. The additional 3D breed loss brings another reduction by a similar margin, which is consistent with the perceptual study.
Again, all pairwise differences are highly significant (paired $t$-test, $p\!<\!0.0001$).
Furthermore, the gains are consistent across breeds: For 19 out of 20 breeds we get the same order, WLDO \textgreater BARC$_\text{nobreed}$ \textgreater BARC$_\text{sim}$ \textgreater BARC$_\text{sim+3D}$.

\begin{table}[!tb]\centering
\begin{center}
\small
\begin{tabular}{ |c|c|c| } 
\hline
Experiment Settings & \multicolumn{2}{|c|}{AMT Results} \\
  & Votes & Percentage \\
\hline
\textcolor{black}
{
$\boldsymbol{L}^\textbf{B}_\textbf{sim}$} vs.\ no breed losses & 638 : 382 &
\textcolor{black}{\textbf{62.55\%}} : 37.45\%\\    %
\textcolor{black}{$\boldsymbol{\{}\boldsymbol{L}^\textbf{B}_\textbf{sim}\,,\,\boldsymbol{L}^\textbf{B}_\textbf{3D}\boldsymbol{\}}$}
vs.\ $L^\text{B}_\text{sim}$ & 670 : 440  & \textcolor{black}
{
\textbf{60.36\%}} : 39.64\% \\     %
\textcolor{black}{
$\boldsymbol{\{}\boldsymbol{L}^\textbf{B}_\textbf{sim}, \boldsymbol{L}^\textbf{B}_\textbf{3D}\boldsymbol{\}}$} vs. WLDO & 998 : 82 & \textcolor{black}{\textbf{92.41\%}} : 7.59\%    \\ %
\hline
\end{tabular}
\vspace{-0.1in}
\caption{\textit{Perceptual Studies.} Ablation of breed losses and comparison with WLDO. See text.}
\label{tab:main_amt_breed_losses_main_paper}
\end{center}
\vspace{-0.1in}
\end{table}

\begin{table}[!tb]\centering
\small
\begin{tabular}{|c|c|c|c|c|}
\hline
Method & WLDO & \multicolumn{3}{|c|}{BARC}\\ \cline{3-5}
& & $\!$no breed loss$\!$ & $L^\text{B}_\text{sim}$ & \{$L^\text{B}_\text{sim}$, $L^\text{B}_\text{3D}$\} \\ \hline
Error [m] & 0.1155 & 0.0858 & $\,\,$0.0776$\,\,$ & \textbf{0.0695}  %

\\\hline                
\end{tabular}
\vspace{-0.1in}
\caption{\textit{3D Shape Evaluation.} Average over  20 breeds.}
\label{tab:quant_toy_eval}
\end{table}

\noindent \textbf{Breed Similarity Loss:} So far we have considered the two parts of the breed similarity loss $L_\text{sim}^B$ in conjunction. To show the separate contributions of breed classification ($L_\text{cs}^B$) and breed triplet affinity ($L_\text{triplet}^B$) we evaluate consistency with a breed prototype using different weights for the two terms, Tab.~\ref{tab:ablation}. $w_{cs}^{B}$, $w_{triplet}^{B}$, $w_{3D}^{B}$ denote the weights for the classification, triplet and 3D-CG model loss, respectively. All other loss terms (regularizers, reprojection losses) remain fixed. Too a high weight on 3D shape similarity degrades the fit to the 2D image evidence (IoU, PCK). A good trade-off is $w_\text{triplet}^{B}=5$.

\begin{table}[h]\centering   %
\resizebox{0.98\columnwidth}{!}{%
\begin{tabular}{|cc|c|c|c|c|c|c|}
\hline
\multicolumn{1}{|c|}{Loss weights} & $w_{cs}^{B}$      & -      & 1      & 1      & 1      & 1      & 1      \\ \cline{2-8} 
\multicolumn{1}{|c|}{}                              & $w_{triplet}^{B}$ & -      & -      & 5      & 10      & 5     & 10     \\ \cline{2-8} 
\multicolumn{1}{|c|}{}                              & $w_{3D}^{B}$      & -      & -      & -      & -      & 1      & 1      \\ \hline
\multicolumn{2}{|c|}{Error [m]}  & 0.0858 & 0.0799 & 0.0776 & 0.0776  & 0.0695 & \textbf{0.0694} \\ \hline  %
\end{tabular}
}
\vspace{-0.1in}
\caption{\textit{Ablation study.} Breed prototype consistency error for different setting}
\label{tab:ablation}
\end{table}

 To make the influence of the breed information more tangible, we also visualize the effect of the breed similarity loss. Figure \ref{fig:latent} shows a t-SNE visualization of the latent feature spaces learned by (left) a network without $L^\text{B}_\text{sim}$ and (middle, right) an identical network trained with $L^\text{B}_\text{sim}$.
 The breed similarity pulls dogs of the same breed closer together in the latent space $z$, which is closely linked to the body shape parameters $\beta$. Different saturation levels of the same color indicate breeds within the clade. Even though the notion of clades is not imposed or made explicit anywhere in our network, breeds of the same clade tend to cluster.
 This suggests that not only within breeds, but also above breed level, shape knowledge can be transferred.

\section{Conclusion}

We present a method to reconstruct 3D pose and shape of dogs from images. Monocular 3D reconstruction of articulated objects is an unconstrained problem that requires strong priors on 3D shape and pose. We overcome the limitation of current 3D shape models of animals by training for model-based shape prediction with a novel breed-aware loss. %
We obtain state-of-the-art estimates of 3D dog shape and pose from images while also producing consistent, breed-specific 3D shape reconstructions.
Our results outperform previous work metrically and perceptually. %
Combining visual appearance and genetic information through breed labels, we obtain a latent space that expresses relations between different breeds in accordance with recent studies on the dog breed genome. We believe this is the first work that combines breed information for learning to reconstruct 3D animal shape, and we hope it will be the basis of further investigation for other species.

\noindent \textbf{Limitations and Ethics.}
BARC is limited by its shape space and is not able to go outside it.
Given the high-quality regression results, future work should explore learning an improved shape space from images by exploiting breed constraints.
We focused mainly on shape, but pose and motion are also important, and learning models of these from image data may be possible using our methods.
Our research uses public image sources of dogs, and no animal experiments were conducted.
While we focus on dogs, our method should be applicable to other animals and may eventually find positive uses in conservation, animal science, veterinary medicine, \etc.

\noindent \textbf{Aknowledgements.} This research was supported by the Max Planck ETH Center for Learning Systems. A conflict of interest disclosure for Michael J.\ Black can be found here \url{https://files.is.tue.mpg.de/black/CoI_CVPR_2022.txt}.

{
    \clearpage
    \small
    \bibliographystyle{ieee_fullname}
    \bibliography{references}      %
}

\newpage
\beginsupplement

\twocolumn[{%
  \renewcommand\twocolumn[1][]{#1}%
  \vspace{1cm}
  \begin{center}
  { \Large \textbf{BARC: Learning to Regress 3D Dog Shape from Images by Exploiting Breed \\  
  \vspace{0.3cm}
  Information -- Supplementary Material}}
  \end{center}
  \vspace{1.7cm}
}]

\section{AMT Perceptual Studies}
\begin{figure}[b]
    \centering
        \includegraphics[width=0.95\linewidth]{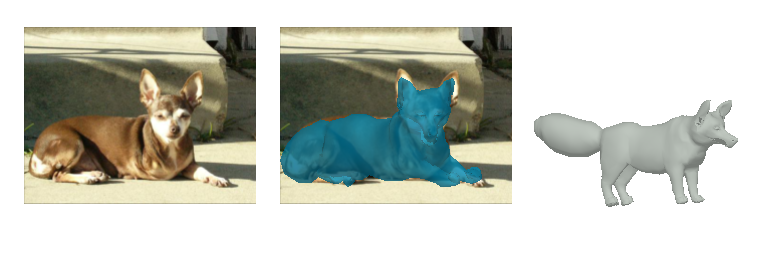}
    \caption{\textit{Misleading reprojection errors} Both IoU and PCK are sometimes misleading, as they can be high for poor 3D estimates.}
    \label{fig:iou_misleading}
\end{figure}

3D shape evaluation based on 2D reprojection errors can be misleading. Figure \ref{fig:iou_misleading} shows an example where the IoU score is high, but the estimated 3D shape of the dog not accurate.
In order to better evaluate predicted shapes in 3D, we propose an evaluation based on breed prototype consistency as well as perceptual studies. While results of all evaluation methods are shown in the main paper, we elaborate here more on our procedure to perform perceptual studies.
\begin{figure}[t]
  \centering
  \includegraphics[width=1\linewidth]{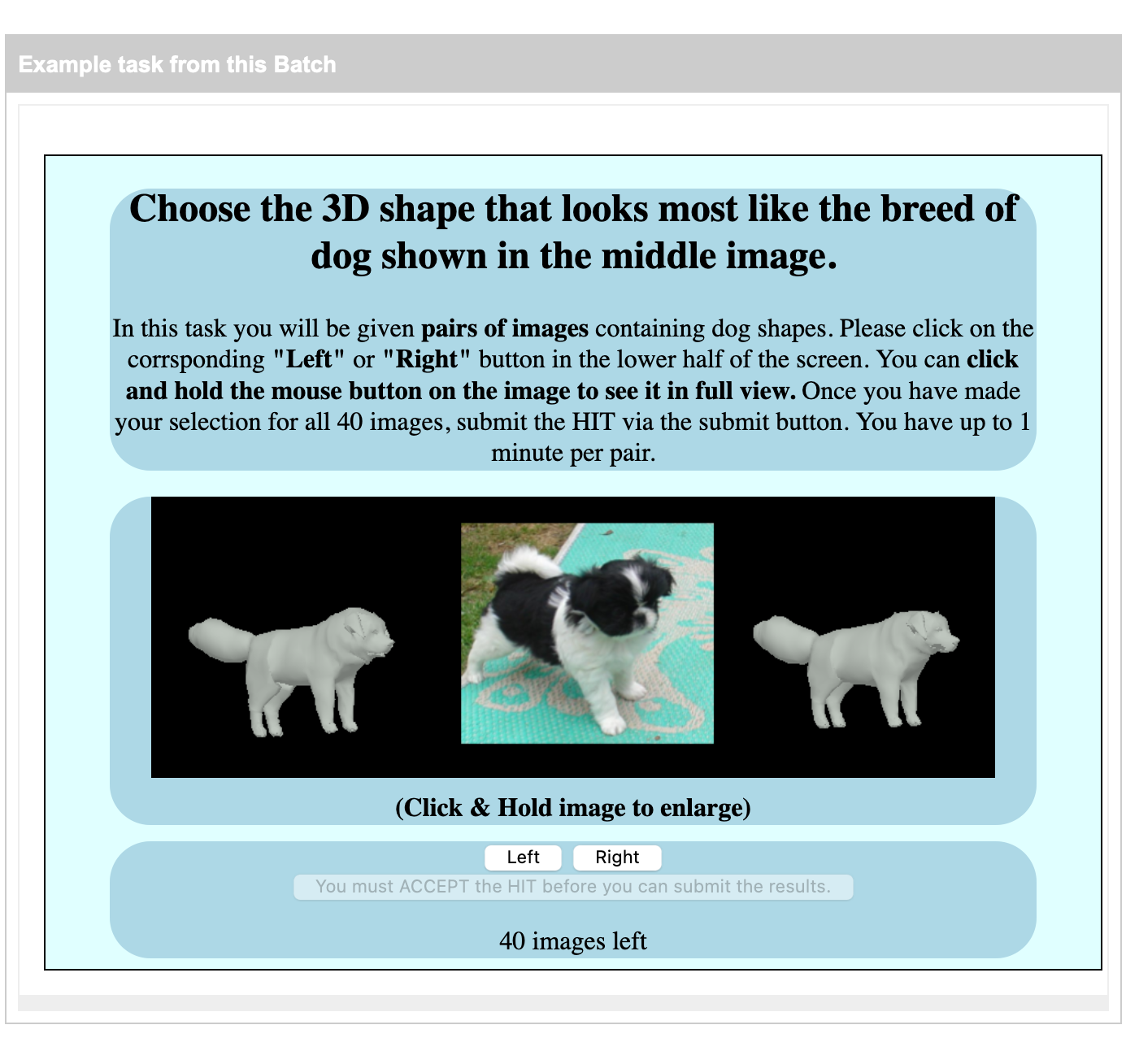}
  \vspace{-0.2in}
  \caption{\textit{AMT Framework.} The picture shows an example screenshot from the perceptual studies that we ran on Amazon Mechanical Turk. }
\label{fig:amt_framework}
\end{figure}
Controlled perceptual tasks are designed to evaluate our method relative to (1) the SOTA or (2) to an ablated model.
Workers on Amazon Mechanical Turk (AMT) judge which of two rendered 3D body shapes better fits a query dog image. 
Figure \ref{fig:amt_framework} shows the framework that we provide to the AMT workers.
We show each worker an image that contains a dog, our predicted 3D model in T-pose and the model in T-pose from SOTA or ablated method. We do not present the predicted 3D posed models in order to focus workers on shape. The left-right ordering of the rendered meshes is random.
We let each worker first process 8 samples to get used to the task and then use the next 30 hits. The task is split in 4 batches with 30 samples each. We have 10 workers for each batch. This gives us a total of 1200 hits. In  order  to  verify the  workers  understand the  task  and perform  it  diligently, we include  two catch  trials  in  each batch.  These are extreme cases where one 3D shape is so far off that only one answer is plausible. For all quantitative results reported, votes from workers who failed one or both catch trials are ignored.

\section{3D CG Models}
We propose to use 3D CG models to help training our network, in case such models are available. See Tab. \ref{tab:cg_models} for a list of 3D CG models and corresponding breeds which BARC uses in its 3D model loss.
\begin{table}[h]\centering
\begin{tabular}{|l|l|}
\hline
\multicolumn{1}{|c|}{\textbf{Breed}}                                      & \multicolumn{1}{c|}{\textbf{Stanford Extra Name}}                                     \\ \hline
\begin{tabular}[c]{@{}l@{}}American Staffordshire \\ Terrier\end{tabular} & \begin{tabular}[c]{@{}l@{}}n02093428-American\_\\ Staffordshire\_terrier\end{tabular} \\ \hline
Boxer                                                                     & n02108089-boxer                                                                       \\ \hline
German Shepherd                                                           & n02106662-German\_shepherd                                                            \\ \hline
Doberman                                                                  & n02107142-Doberman                                                                    \\ \hline
\begin{tabular}[c]{@{}l@{}}Staffordshire \\ Bullterrier\end{tabular}      & \begin{tabular}[c]{@{}l@{}}n02093256-Staffordshire\\ \_bullterrier\end{tabular}       \\ \hline
French Bulldog                                                            & n02108915-French\_bulldog                                                             \\ \hline
Bull Mastiff                                                              & n02108422-bull\_mastiff                                                               \\ \hline
Great Dane                                                                & n02109047-Great\_Dane                                                                 \\ \hline
Italian Greyhound                                                         & n02091032-Italian\_greyhound                                                          \\ \hline
Rottweiler                                                                & n02106550-Rottweiler                                                                  \\ \hline
Siberian Husky                                                            & n02110185-Siberian\_husky                                                             \\ \hline
\end{tabular}
\caption{\textit{3D CG models.} Models used for our 3D model loss $L_{3D}^B$}
\label{tab:cg_models}
\end{table}

\section{Keypoint Weights}
Table \ref{tab:kp_weights} shows for each keypoint the weight that was used as part of the weighted keypoint loss.  

\begin{table}[h!]\centering
\begin{tabular}{|l|l|l|l|}
\hline
keypoint                & w & keypoint            & w \\ \hline
left front leg, paw     & 3 & right rear leg, top & 2 \\
left front leg, middle  & 2 & tail start          & 3 \\
left front leg, top     & 2 & tail end            & 3 \\
left rear leg, paw      & 3 & base left ear       & 2 \\
left rear leg, middle   & 2 & base right ear      & 2 \\
left rear leg, top      & 2 & nose                & 3 \\
right front leg, paw    & 3 & chin                & 1 \\
right front leg, middle & 2 & left ear tip        & 2 \\
right front leg, top    & 2 & right ear tip       & 2 \\
right rear leg, paw     & 3 & left eye            & 1 \\
right rear leg, middle  & 2 & right eye           & 1 \\ \hline
\end{tabular}
\caption{\textit{Keypoint weights.} Weights that are used within the weighted keypoint loss.}
\label{tab:kp_weights}
\end{table}

\section{Failure Case Analysis}
We divide the failure cases in two main groups: shape and pose failures.

\noindent\textbf{Pose Failure Cases:} At development time we have trained our network with various pose priors, such as a mixture of gaussians prior as in \cite{zuffi2018lions, biggs2020wldo}, a variational auto-encoder as in \cite{zuffi2019three} and our final normalizing flow pose prior. One failure mode that goes through all priors is the erroneous prediction of dogs not facing the camera. The Stanford Extra training set is unbalanced in the sense that it shows many dogs from a front- or side-view. Furthermore, most of the dogs do not bend the front legs as they are either sitting, laying or standing, this leads to challenges when predicting poses for dogs with heavily bent wrists. As training with different pose priors lead to similar error cases, we believe that those challenges are not structural problems of the pose prior, but rather of the image dataset. Nevertheless, it might be worth examining different training schedules such that rare poses obtain higher weights or are repeated more often.  
One more thing worth mentioning is, that often perceived 3D quality from front view is considerably higher than from side-views. A strong 3D regularization is inevitable.
Predictions for laying and sitting dogs could be improved by training a pose prior on a more suitable 3D pose dataset.
Furthermore, BARC has troubles predicting poses for dogs that are only partly visible. 

\noindent \textbf{Shape Failure Cases:} Our breed losses help to regularize dog shape. BARC can predict more reasonable shapes, especially for dogs that are not fully visible from the side. Never the less, we do sometimes observe shortened limbs when they are difficult to predict due to poses such as a dog laying and facing the camera.  As discussed in the main paper, working with a single shape for each dog breed is not an option, as there is not negligible intra-class variability. Another challenge is dog hair. First, shape variability can become enormous, consider for example differently sheared poodles. Secondly, long hair does swing and the shape that we want to predict for a dog with fluffy hair is not clearly defined. In such cases, representing a dog with a mesh is not ideal. 

\noindent \textbf{Some Visual Examples of Failure Cases:}
We show four failure cases in Figure \ref{fig:failure_cases}: 
\begin{figure}[b]
\centering
\includegraphics[width=0.8\linewidth]{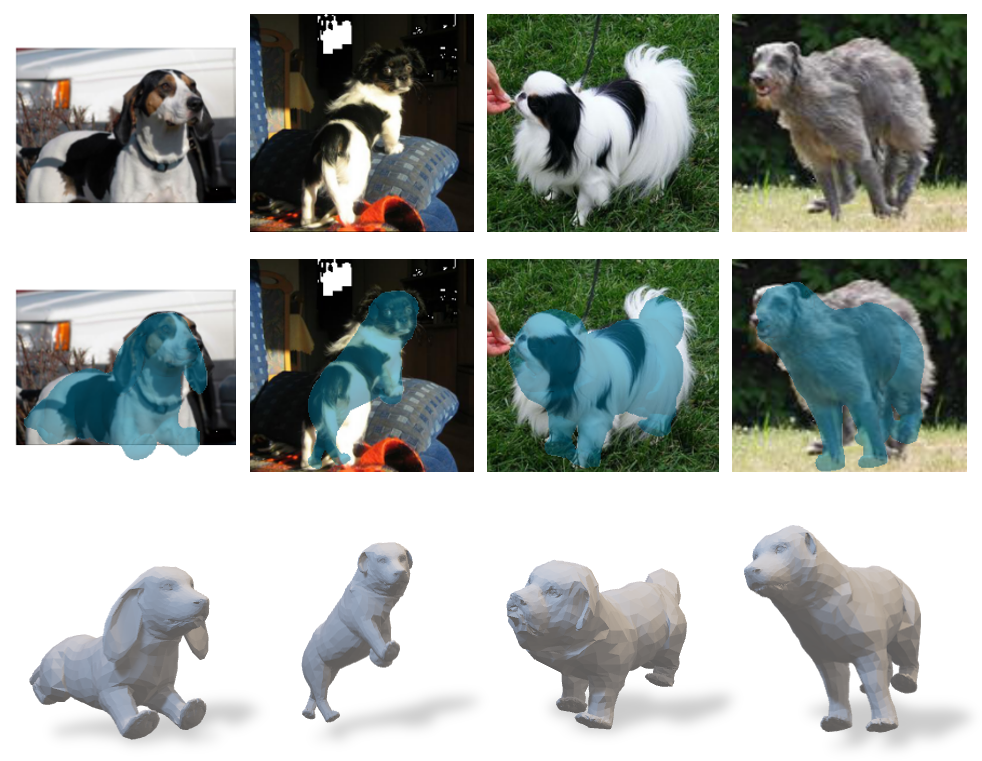}
\caption{\textit{Failure Cases.}Pose and shape failure cases.}
\label{fig:failure_cases}
\end{figure}
(1) a dog which is not fully visible, our prediction shows a shrunken body. (2) most training images show dogs that face the camera. When the dog is turned away, pose prediction fails. (3) a Japanese Spaniel with lots of hair. Shape prediction for such breeds is difficult. (4) A dog that is hard to recognize and where, in part, the difficult pose is compensated by a wrong shape – instead of bending the back, the dog is given a stouter body.

\section{Qualitative Results}
\begin{figure*}[!tb]
\centering
\includegraphics[width=0.6\linewidth]{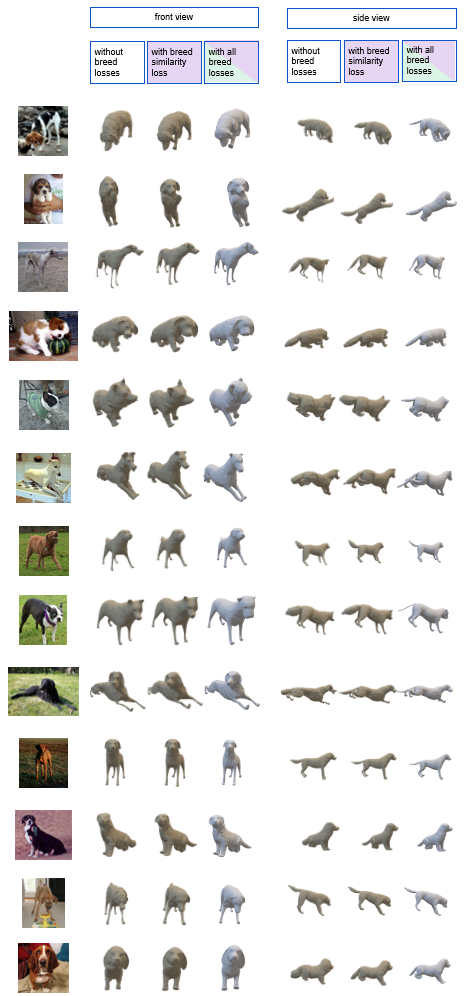}
\caption{\textit{Ablation Study.} Qualitative comparison of from left to right (1) our method trained without any breed losses (2) our method trained with similarity breed loss only (3) BARC (our method). We show for various input images front views as well as side views.}
\label{fig:supmat_qualitative_ablation_study}
\end{figure*}

\begin{figure*}[!tb]
\centering
\includegraphics[width=0.8\linewidth]{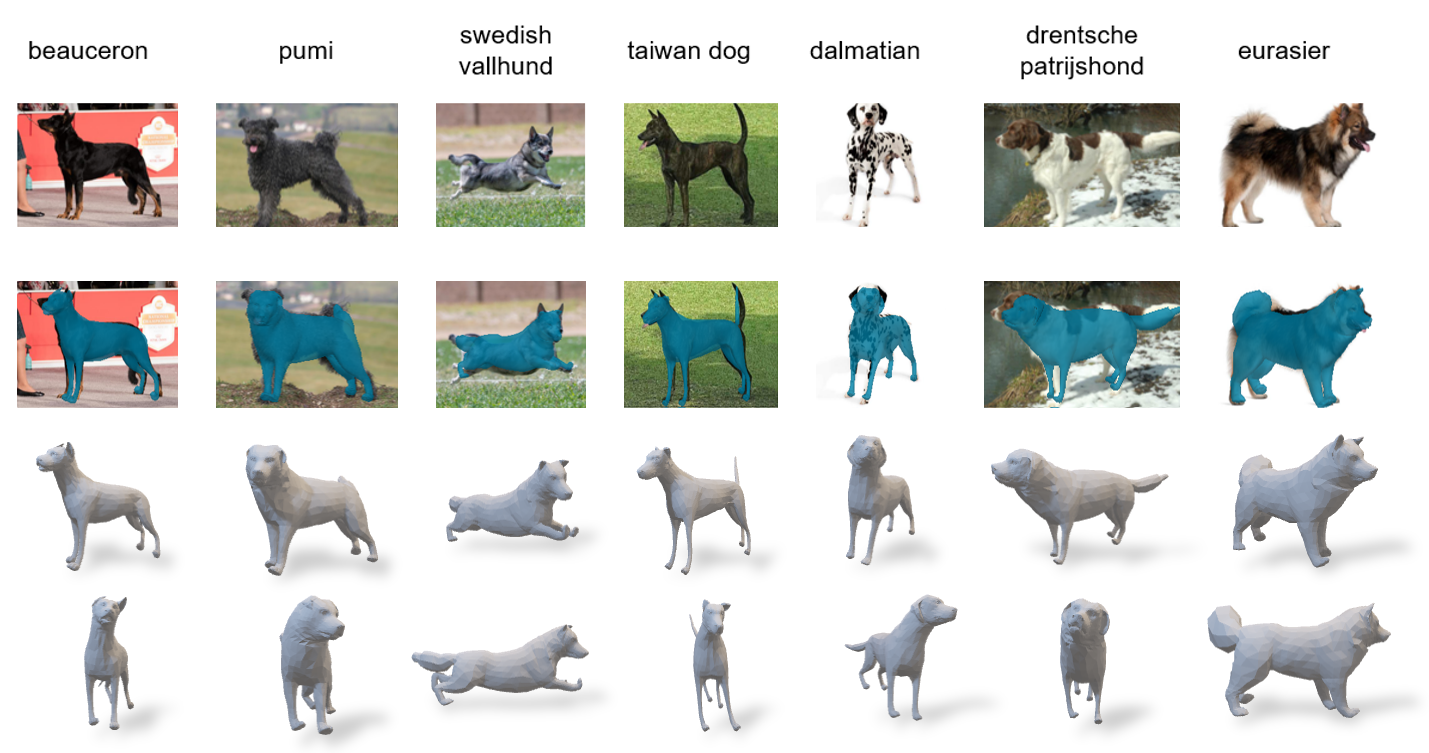}
\caption{\textit{Results for Unseen Breeds.} Qualitative results of BARC (our method) on images of previously unseen breeds. All test images are downloaded from the American Kennel Club web page. We show for various input images an overlay, front view as well as side view of our predicted dog.}
\label{fig:supmat_qualitative_AKC}
\end{figure*}

\begin{figure*}[!b]
    \centering
        \includegraphics[width=0.65\linewidth]{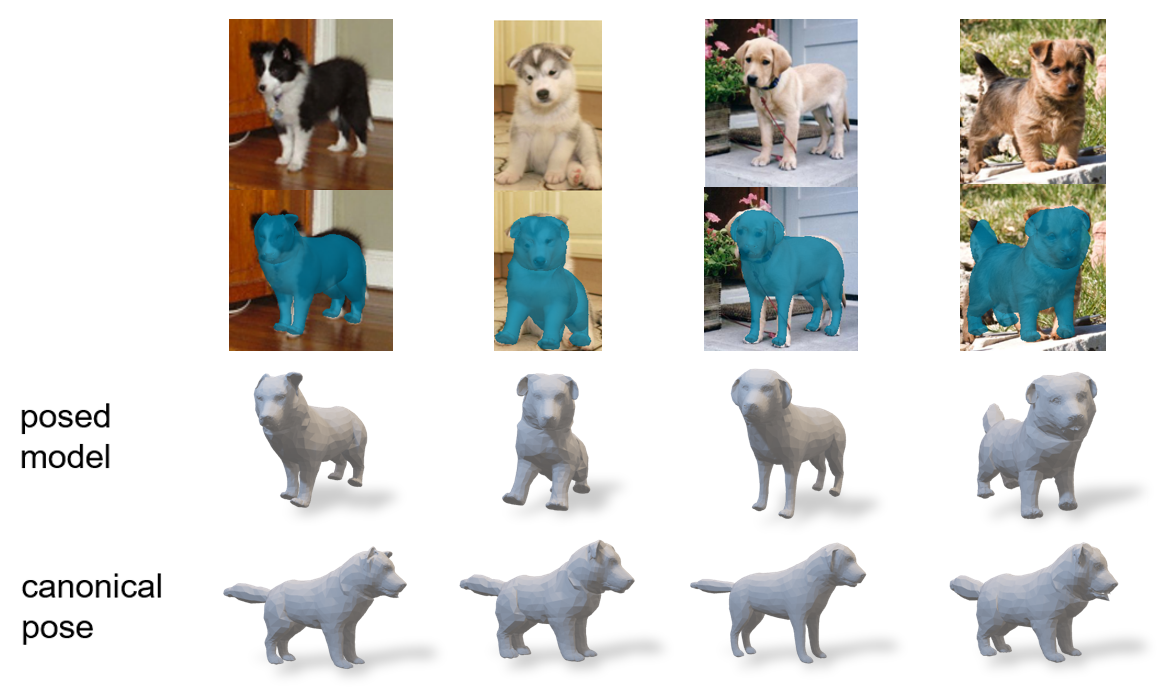}
    \caption{\textit{Puppies.} Qualitative results on puppies from the Stanford Extra test set.}
    \label{fig:puppies}
\end{figure*}

\begin{figure*}[!b]
    \centering
        \includegraphics[width=0.90\linewidth]{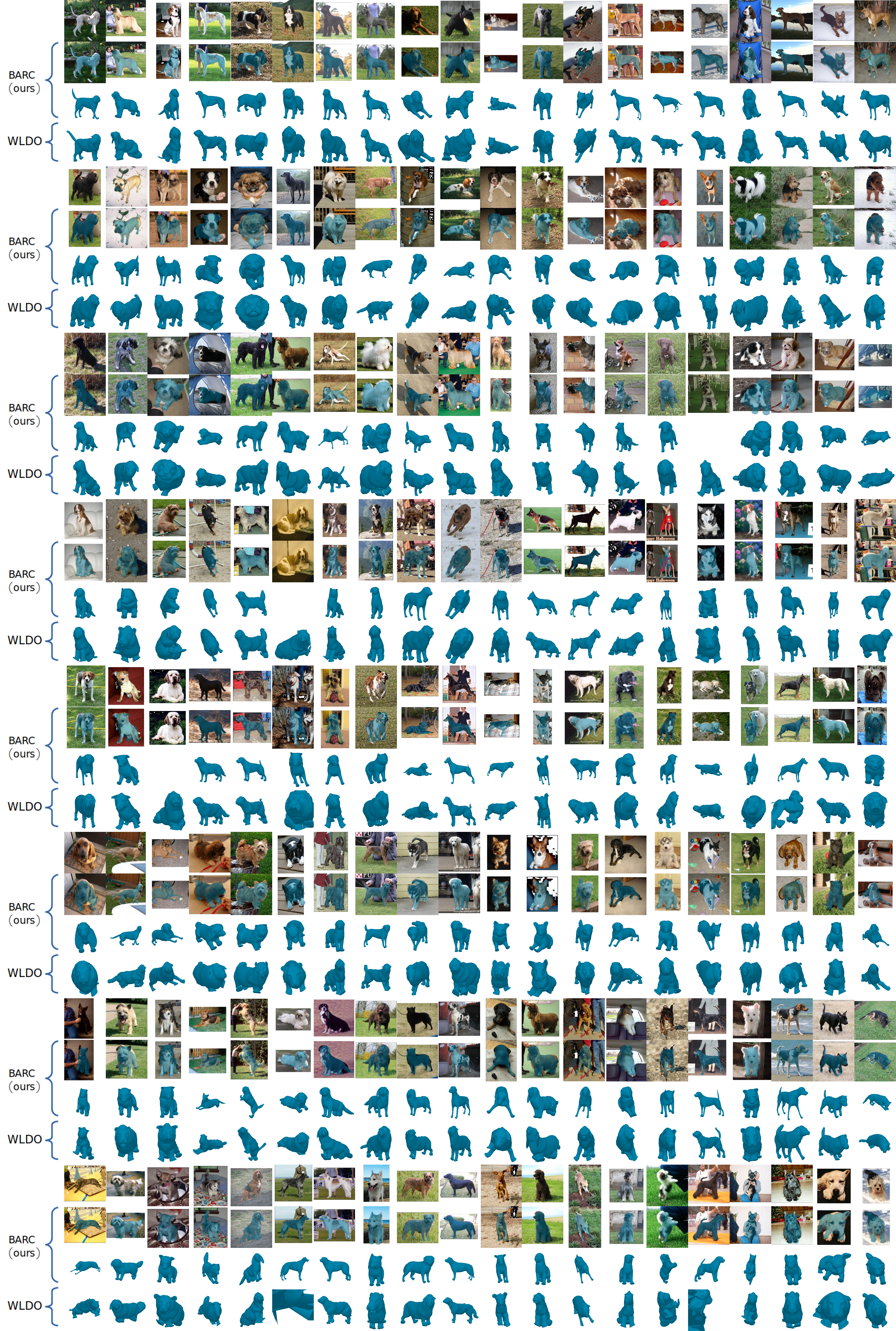}
    \caption{\textit{Randomly sampled results.} We show qualitative results on the Stanford Extra test set: for each sample an input image, the overlay of our prediction (BARC) with that image, our prediction and previous state-of-the-art (WLDO).}
    \label{fig:random_res}
\end{figure*}

In this section we present additional qualitative results.
Figure \ref{fig:supmat_qualitative_ablation_study} shows results for ablated versions of BARC. To the left we render results from our method without any of the breed related losses, in the middle results with the breed similarity loss only and to the right with the breed similarity loss as well as the 3D CG model loss. For each of the three versions we show front as well as a side view. 
Finally, we test BARC on images of previously unseen breeds. All of those images are downloaded from the American Kennel Club web page. Figure \ref{fig:supmat_qualitative_AKC} illustrates an overlay of our prediction on the input image as well as front and side view for each of the seven dogs. We observe that BARC can generalize well to new breeds. Furthermore it generalizes to puppies, as illustrated in Figure \ref{fig:puppies}. 
For the figures in the paper, we select results to illustrate variety. Last, in Figure \ref{fig:random_res} we present results on completely randomly sampled Stanford Extra test set images. For each input image we show the overlap of our prediction with this image, a 3D visualization of our prediction and a 3D visualization of the previous state-of-the-art method WLDO.

\end{document}